\documentclass[conference]{IEEEtran}
\usepackage{cite}
\usepackage{amsmath,amssymb,amsfonts}
\usepackage{wrapfig}
\usepackage{color}
\usepackage[hyperfootnotes=false]{hyperref}
\usepackage{cleveref}
\usepackage{algorithmic}
\usepackage{graphicx}
\usepackage{textcomp}
\usepackage[dvipsnames]{xcolor}
\def\BibTeX{{\rm B\kern-.05em{\sc i\kern-.025em b}\kern-.08em
    T\kern-.1667em\lower.7ex\hbox{E}\kern-.125emX}}
\usepackage{enumitem}
\usepackage{mathtools}
\usepackage{booktabs}
\usepackage{subfigure}
\usepackage{footmisc}
\usepackage{etoolbox}
\makeatletter
\patchcmd{\@makecaption}
  {\scshape}
  {}
  {}
  {}
\makeatletter
\patchcmd{\@makecaption}
  {\\}
  {.\ }
  {}
  {}
\makeatother

\Crefname{equation}{Eq.}{Eqs.}
\Crefname{section}{Sect.}{Sects.}
\Crefname{figure}{Fig.}{Figs.}
\Crefname{tabular}{Tab.}{Tabs.}
\Crefname{figure}{Fig.}{Figs.}
\Crefname{tabular}{Tab.}{Tabs.}

\DeclareMathOperator*{\argmin}{argmin} % no space, limits underneath in displays
\DeclareMathOperator*{\argmax}{argmax} % no space, limits underneath in displays

\DeclareFontFamily{OT1}{pzc}{}
\DeclareFontShape{OT1}{pzc}{m}{it}{<-> s * [1.100] pzcmi7t}{}
\DeclareMathAlphabet{\mathpzc}{OT1}{pzc}{m}{it}

\newcommand{\varA}[1]{{\operatorname{#1}}}

\begin{document}
\title{Periodic Extrapolative Generalisation in Neural Networks}

\author{
    \IEEEauthorblockN{Peter Belcak}
    \IEEEauthorblockA{
    \textit{ETH Z\"urich}\\
    Z\"urich, Switzerland \\
    belcak@ethz.ch}
    \and
    \IEEEauthorblockN{Roger Wattenhofer}
    \IEEEauthorblockA{
    \textit{ETH Z\"urich}\\
    Z\"urich, Switzerland \\
    wattenhofer@ethz.ch}
}

\maketitle

\begin{abstract}
% The learning of the simplest possible computational pattern -- periodicity -- is a current frontier in neural network research, aiming to extend the neural generalisation ability beyond the learning domain.
% To achieve this goal would mean enabling deep neural networks to recognize non-conditional patterns and consequently infer complex representations indefinitely extending beyond the learning data, greatly aiding the effort towards general artificial intelligence.

% We systematically investigate the generalisation abilities of both classical and recently proposed architectures on a set of benchmarking tasks we introduce and find that traditional sequential models still outperform the novel architectures designed specifically for extrapolation.
% We make our benchmarking and evaluation toolkit, \textsc{PerKit}, easily accessible and freely available to facilitate future work in the area. 

The learning of the simplest possible computational pattern -- periodicity -- is an open problem in the research of strong generalisation in neural networks.
We formalise the problem of extrapolative generalisation for periodic signals and systematically investigate the generalisation abilities of classical, population-based, and recently proposed periodic architectures on a set of benchmarking tasks.
We find that periodic and ``snake'' activation functions consistently fail at periodic extrapolation, regardless of the trainability of their periodicity parameters.
Further, our results show that traditional sequential models still outperform the novel architectures designed specifically for extrapolation, and that these are in turn trumped by population-based training.
We make our benchmarking and evaluation toolkit, \textsc{PerKit}\footnote{\textsc{PerKit}: A toolkit for the study of periodicity in neural networks. Available at \url{https://github.com/pbelcak/perkit}\,.\label{footnote:repo_link}}, available and easily accessible to facilitate future work in the area. 

\end{abstract}


\section{Introduction}
Neural networks have for long been hailed for their ability to construct representations capturing the prevailing patterns in data provided, and to meaningfully generalise to previously unseen data of similar nature.
The ability of a network to perform \textit{interpolating} generalisations is commonly associated with settling at loss surface minima that are surrounded by large areas of flatness \cite{foret2020sharpness}.
This is also the mode of generalisation exploited by autoencoder \cite{hinton2006reducing, goodfellow2016deep} and variational autoencoder \cite{higgins2016beta} architectures.
Generalising with neural networks by \textit{extrapolation} is often impossible for non-linear tasks, and only sometimes possible if the appropriate non-linearities are encoded in the architecture and input representation \cite{xu2020neural}, or if the task is fundamentally algorithmic and tailored to that end \cite{d2022deep,belcak2022fact}.

We focus on the extrapolative generalisation and observe that the simplest possible patterns permitting extrapolation are periodic, requiring only a single-tape write-only Turing machine unconditionally outputting the pattern without halting.
In the context of neural networks, we distinguish between three distinct periodicity-related learning tasks, namely the learning of a periodic signal
\begin{enumerate}
    \item[L1.] \label{item:noExtrapolationJustSignal} without the need for extrapolation (e.g. for conditional generation),
    \item[L2.] \label{item:extrapolationNoPeriod} with the need for extrapolation but with the knowledge of the (approximate) period a priori, and
    \item[L3.] \label{item:extrapolationAndPeriod} with the need for extrapolation but with period being a learned parameter.
\end{enumerate}
Naturally, even more complex tasks (more difficult than L3) can be considered, for example the learning to separate individual periodic components from which a class of signal is composed.
These are beyond the scope of our work.

We show that models succeeding in L1 are prone to fail at L2, and that models proposed for L2 fail or struggle with L3.
This leads to an establishment of an order of learning difficulty for neural networks.
Given a downstream task, all models of higher levels generally succeed at lower levels of the hierarchy.
We further find that the standard evaluation metrics used for regression tasks do not represent the relative successes of models learning periodic patterns well when training models for extrapolation.
After classifying the common prediction faults for L2-L3 tasks, we therefore propose new metrics for this purpose. 

Having noticed that the current literature lacks a unified set of criteria to evaluate the suitability of models for L1-L3 tasks, we develop a benchmarking dataset for periodic generalisation and produce a comprehensive comparative study of models proposed so far.
On top of models that have been introduced specifically to tackle the problems of learning periodic functions and periodic generalisation, we also evaluate classical and evolutionary population-based training (PBT) methods on the same benchmarking dataset.
These methods are consistent in their execution with their peers in literature and their evaluation results give a natural insight into what can be achieved by making repeated ``informed guesses'' about the periods of the signals being learned.
Since PBT methods operate on a population of simple networks that are iteratively trained and adjusted for best fit with the target signal, their performance serves as an expectation baseline for future work aiming to address periodic generalisation in a more sophisticated manner.

% Our contributions are (i) the formal specification of the problem of periodic extrapolative generalisation with neural networks, (ii) a systematic evaluation of the behaviour of the models recently proposed for learning of periodic signals with neural networks, and (iii) the introduction of a benchmarking toolkit consisting of a (iiia) dataset and tasks tailored to meaningfully asses the ability of models to perform this mode of generalisation and (iiib) a unified framework designed to accelerate further research in the field.
Our contributions are therefore
\begin{itemize}
    \item the formal specification of the problem of periodic extrapolative generalisation in neural networks and the establishment of difficulty hierarchy L1-L3 (\Cref{section:generalisationOverview}),
    
    \item a systematic evaluation of the behaviour of the models recently proposed for learning of periodic signals with neural networks (\Cref{section:evaluation}),
    
    \item the establishment of an expectation baseline on the possible performance of neural models for the periodic extrapolation tasks by performing population-based training of simple neural networks (\Cref{section:pbt} and \Cref{section:evaluation}), and
    
    \item the introduction of a comprehensive, accessible benchmarking toolkit, consisting of a dataset and tasks tailored to meaningfully asses the ability of models to perform periodic extrapolative generalisations and of a unified software framework designed to accelerate further research in the field (Footnote \ref{footnote:repo_link}).
\end{itemize}

%%%%%%%%%%%%%%%%%%%%%%%%%%%%%%%%%%%%%%%%%%%%%%%%%%%%%%%%%%%%%%%%%%%%%%%%%%%%%%%%%%%%%%%%%%%%%%%%%%%%%%%%%%%%%%%%%%%%%%%%%%%%%%%%%
\section{Related Work}
\label{section:relatedWork}
%%%%%%%%%%%%%%%%%%%%%%%%%%%%%%%%%%%%%%%%%%%%%%%%%%%%%%%%%%%%%%%%%%%%%%%%%%%%%%%%%%%%%%%%%%%%%%%%%%%%%%%%%%%%%%%%%%%%%%%%%%%%%%%%%

\textit{Trigonometric Activations.}
A feedforward neural architecture aiming to mimic the behaviour of Fourier series has first been proposed in \cite{gallant1988there}, introducing the ``cosine squasher'' activation, very similar in shape to the now-familiar sigmoid but with domain directly adjusted to $\pi$-periodicity.
A further attempt of similar nature has been made in \cite{silvescu1999fourier}, which introduced an activation featuring frequency-adjusted and frequency-offset cosines, aiming for ``Fourier Neural Networks''.
More recently, \cite{liu2013fourier} introduced an activation based on a linear combination of frequency-adjusted and frequency-offset sine and cosine.
These trigonometric activations have, however, been singled out for causing problems in optimisation due to their non-monotonicity \cite{parascandolo2016taming,zhumekenov2019fourier}.
These proposed approaches are included in our evaluation.

\textit{Monotonicitised Trigonometric Activations.}
To address what has been thought of as the main fault of the work on neural networks mimicking the behaviour of Fourier series, \cite{ziyin2020neural} proposed $x+\sin(x)$, $x+\cos(x)$, and $x+\sin^2(ax)$ activations and demonstrated that they possess some potential for generalisation beyond the training domain while still performing well on standard tasks defined on real-world data such as classifying the MNIST dataset.
In later sections, we show that these activation functions consistently fail at periodic extrapolation, and that they offer little to no improvement over purely periodic activations.

\textit{Fourier-Based Decoder.} \cite{lee2021conditional} designed a VAE-based architecture giving coefficients of Fourier series as decoder output and evaluated it against both synthetic and ECG data, showing superior performance in comparison to methods common in time-series analysis.
The authors, however, normalise all signals to have period $1$ and do not explore extrapolative generalisation behaviour of their model. 

\textit{Recurrent Architectures.} Recurrent architectures are the canonical tool for time-series prediction but have been criticised in \cite{lee2021conditional} for their shortcomings when the input samples are fed in irregular time intervals or contain noisy observations.
We find that they are remarkably robust nevertheless.

\textit{Classical Approaches.} In the context of structure discovery, several methods have been proposed for decomposing time series in an explainable fashion to arrive at a composition of patterns that permits straightforward extrapolation \cite{duvenaud2013structure,duvenaud2014automatic}. While our work carries some resemblance to the motifs of this research, our goal is to investigate the extrapolative abilities of neural networks, rather than to propose neurosymbolic or evolutionary methods to further research in structure discovery.

%%%%%%%%%%%%%%%%%%%%%%%%%%%%%%%%%%%%%%%%%%%%%%%%%%%%%%%%%%%%%%%%%%%%%%%%%%%%%%%%%%%%%%%%%%%%%%%%%%%%%%%%%%%%%%%%%%%%%%%%%%%%%%%%%
\section{Learning for Periodic Extrapolative Generalisation}
\label{section:generalisationOverview}
%%%%%%%%%%%%%%%%%%%%%%%%%%%%%%%%%%%%%%%%%%%%%%%%%%%%%%%%%%%%%%%%%%%%%%%%%%%%%%%%%%%%%%%%%%%%%%%%%%%%%%%%%%%%%%%%%%%%%%%%%%%%%%%%%

\subsection{Formal Specification}
\label{section:modesOfDegradataion}
Let $\mathcal{D}_T \subseteq \mathbb{R}^{d_i} \times \mathbb{R}^{d_o}$ be a training domain (dataset) where $d_i,d_o$ are the input and output dimensions, respectively.
Let $\gamma: \mathbb{R}^{d_i} \to \mathbb{R}^{d_o}$ be a function such that for $\psi \in \mathbb{R}^{d_i}$
\[
    \gamma(\psi k + \phi) = \gamma(\psi (k + 1) + \phi) \, \forall k \in \mathbb{Z}, \phi \in \mathbb{R}^{d_i}
\]
and for $(x, y) \in \mathcal{D}_T$, $y = \gamma(x)$.

The problem of periodic generalisation in neural networks is the problem of designing a neural model $\mathcal{M}$ such that when trained on $\mathcal{D}_T$,
\[
    \mathcal{M}(x) = \gamma(x) \text{ for } x = \psi k_x + \phi_x \not\in \mathcal{D}_T.
\]

Let $\Gamma$ be the set of all pairs $(\gamma,\psi)$ satisfying the above requirement.
Then the dimension of $\text{span}(\psi : (\gamma,\psi) \in \Gamma)$ is \textit{the order of periodicity} of $\mathcal{D_T}$.

It is, however, difficult to find out whether a network outputs a particular value on infinitely many points.
We therefore evaluate the ability of $\mathcal{M}$ to extrapolate periodically by testing its predictions $\mathcal{M}(x)$ against $\gamma(x)$ only for $x \in \mathcal{D}_E$, where $\mathcal{D}_E \subseteq \mathbb{R}^{d_i} \times \mathbb{R}^{d_o}$ with $\mathcal{D}_E \cap \mathcal{D}_T = \emptyset$ is the evaluation domain.

\subsection{Difficulty of Periodicity-Learning Tasks}
\begin{figure}[b!]
    \centering
    \includegraphics[width=0.50\textwidth]{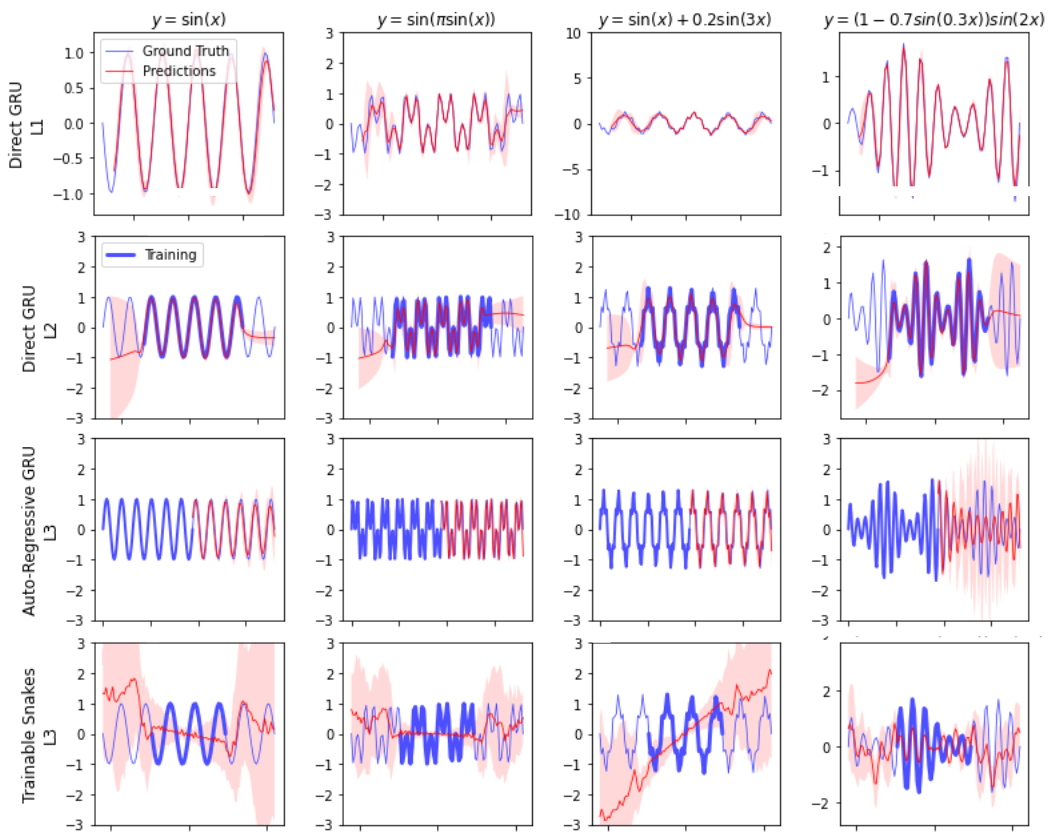}
    \caption{
    An experimental analysis of how a selection of models learns and extrapolates various basic function types.
    The red curve represents the median model prediction and the shaded regions show the 90\% credibility interval from 30 runs.
    GRU networks used 120 units aligned in a single layer, and the snake network was a two-layer feedforward network with 128 snake neurons in the hidden layer.
    }
    \label{fig:hierarchy}
\end{figure}

We illustrate the existence of the difficulty hierarchy L1-L3 of periodicity-learning tasks experimentally.
For the direct GRU and snake experiments, we sample training data randomly from example periodic functions in the range $[-5\pi, 5\pi]$.
In the case of the L1 task, we further add Gaussian noise randomly generated for every epoch of learning with mean $0$ and variance $0.0225$ to the original signal.
For the L2 and L3 tasks we look at the model predictions on the wider range $[-8\pi, 8\pi]$ except for auto-regressive GRU, where we train on $[0\pi, 12\pi]$ and evaluate on $[12\pi, 23\pi]$.
The results are shown in \Cref{fig:hierarchy}.

We see that the GRU recurrent network fed directly with signal time input succeeds in uncovering the original signal despite the presence of noise in training, but fails to generalise the learned signal beyond the training range even when trained on regular inputs without the presence of noise.
We also observe that auto-regressive GRU recurrent networks fare fairly well on the L3 tasks but suffer from loss of information the further away from the training data they extrapolate.
It was presented in \cite{ziyin2020neural} that a snake-activated feedforward neural network can learn to fit a periodic signal if the period is known beforehands.
The bottom row of \Cref{fig:hierarchy} shows that such networks, however, fail to learn periodic signals when the frequency parameter of the snake activation is made trainable, something we elaborate on in 
\Cref{section:snakesInvestigated}.
In our experience, models that succeed at L3 tasks do well on L2 tasks, and those which succeed at L2 tasks mostly do not struggle with L1 tasks either if the period is known.
This illustration is further supported by our results in \Cref{section:evaluation}.

\subsection{Recurrent Extrapolation}
\begin{figure}[b!]
    \centering
    \includegraphics[width=0.50\textwidth]{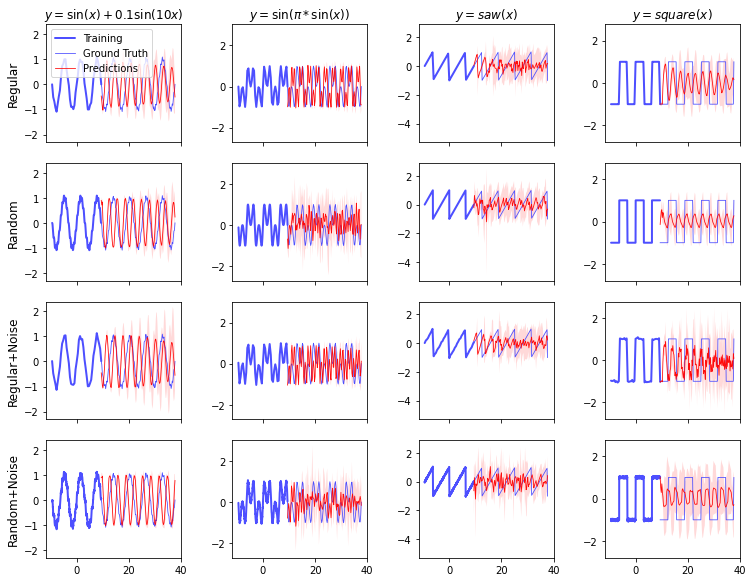}
    \caption{
        The results of a systematic experiment on the ability of auto-regressive RNNs to generalise by extrapolation. 
        Regular sampling means sampling training points at regular intervals, random sampling entails choosing a point at random from equally-spaced bins.
    }
    \label{fig:detailedDecoderChallenge}
\end{figure}

It was previously demonstrated that the extrapolation behaviour of feedforward neural networks with ReLU and $\tanh$ activation functions is dictated by the analytical form of the activation function (ReLU diverges to $\pm\infty$, $\tanh$ tends towards a constant value), and that this result also holds for sigmoidal networks and the corresponding common variants \cite{ziyin2020neural}.
This is despite the fact that feedforward neural networks regularly show excellent performance in approximating sampled functions on training intervals, even in the presence of balanced noise.
We extended on this by looking at the extrapolative properties of recurrent architectures and noted that while recurrent networks in auto-regressive predictive setup succeed in extrapolating reasonably well beyond the training range, they do not do so if the immediate signal past does not serve as a reliable clue to the future.
The results of an experiment showcasing the extrapolative behaviour of RNNs in auto-regressive configuration are depicted in \Cref{fig:detailedDecoderChallenge}

\subsection{Snake Extrapolation}
\label{section:snakesInvestigated}
\begin{figure}[b!]
\vspace{-6pt}
\centering
\begin{subfigure}{}
    \vspace{-22pt}
    \centering
    \includegraphics[width=1\linewidth]{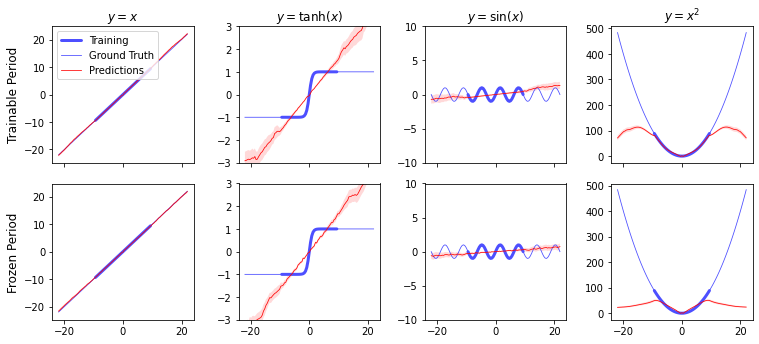}
\end{subfigure}

\begin{subfigure}{}
    \centering
   \includegraphics[width=1\linewidth]{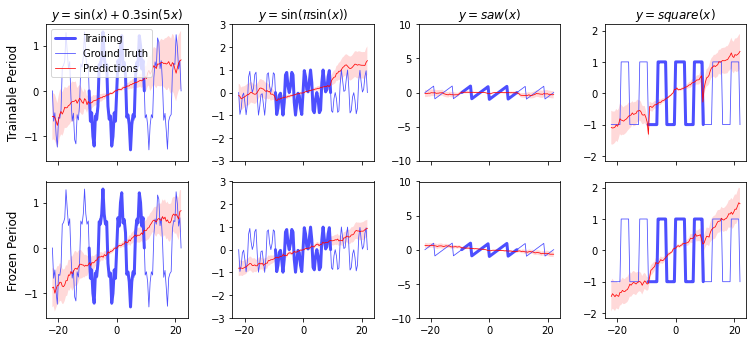}
\end{subfigure}

\caption[Two numerical solutions]{
    A depiction of the tendency of snake neural networks to increase frequency and treat the snake activation as almost-linear component to interpolate the training data.
    \textit{Top.} The results of 40 training instances of snake-powered two-layer feedforward neural networks on general signals.
    The hidden layer is 128 neurons wide and its initial frequencies are drawn according to a continuous uniform distribution on the closed interval $[1,6]$.
    The shaded region represents the 90\% confidence interval, and the median line is bold red.
    \textit{Bottom.} As for the above but on periodic signals.}
    \label{fig:snakesAreLinear}
\end{figure}

\begin{figure}[t]
    \centering
    \includegraphics[width=0.25\textwidth]{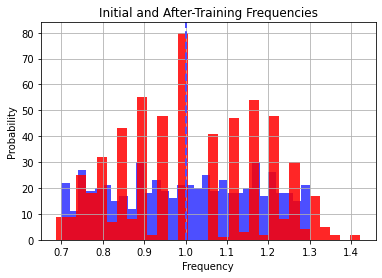}
    \caption{
        The results of an experiment using a single-snake-neuron network to learn an underlying snake signal with parameter $a = 1$.
        The parameter $a$ of the network attempting to learn the signal is trainable and initialised according to a uniform distribution on $[0.7, 1.3]$.
        The blue and red histograms chart the initial (pre-training) and post-training values of parameter $a$ in the network.
        We observe that the trained parameters settle consistently around nearby local minima and only a fractional minority finds the true frequency of $1$.
        }
    \label{fig:snakesVsSnakes}
\end{figure}

To address the shortcomings of standard activation functions in extrapolation, \cite{ziyin2020neural} proposed the family of ``snake'' activation functions.
If the frequency $a$ of the periodic signal to be approximated is known, a two-layer feedforward network with the corresponding snake activation $x+sin^2(ax)$ has been demonstrated to be able to approximately learn the amplitude of the signal.
Authors further proposed to make the parameter $a$ of the activation a learnable parameter and appealed to Fourier convergence for justification of general learnability properties.
Our experiments following the setup and training from \cite{ziyin2020neural} show that regardless of the trainability of the frequency parameter, a snake-activated feedforward neural network often resorts to maximising the frequency and ignores the loss minima corresponding to the Fourier coefficients  (\Cref{fig:snakesAreLinear}).
Further, we observed (\Cref{fig:snakesVsSnakes}) that when the frequency parameter is trainable, such feedforward networks even fail to learn the activation function itself.

\subsection{Common Modes of Signal Degradation in Periodic Extrapolation}
\label{section:modesOfDegradation}
\begin{figure}[t]
    \centering
    \includegraphics[width=0.46\textwidth]{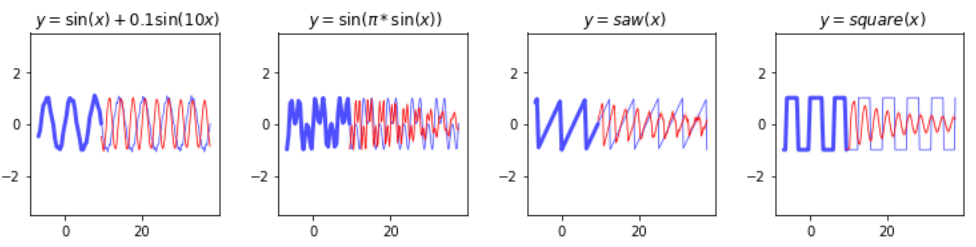}
    \caption{
        An illustration of information loss by recurrent neural networks in extrapolative generalisation.
        The left-most sinusoidal signal is shifted and its frequency is higher than that of the ground truth (periodic speedup).
        Other three signals suffer from slow dissolution of shape and amplitude information.
        In the right-most plot, the recurrent network failed to learn the shape of the signal.
    }
    \label{fig:faults}
\end{figure}

Three types information can be lost or slowly degrading with increasing distance from the training domain, namely the values of frequency and amplitude parameters, and the shape of the underlying periodic function (\Cref{fig:faults}).
In \Cref{fig:detailedDecoderChallenge} we have observed on recurrent networks that these three modes of information loss do not necessarily have to happen at same the pace.

If the data in the evaluation domain $\mathcal{D}_E$ consists of only sampled points, it might be difficult or impossible to judge the quality of predictions when the signal shifts in ``time'' (the argument of $\gamma$), even if only slightly.

In our experimentation, we recognised three modes of model prediction degradation related to the periodicity parameters of the ground-truth $\gamma$:
\begin{itemize}
    \item \textbf{Periodic Shift (SH).} The model clips away or fills in parts of the learned or previously predicted signal periodically in discrete increments, thus making the predictions for individual periods increasingly more shifted with every iteration. 
    
    \item \textbf{Periodic Speedup (SP).} The model believes that the frequency of $\gamma$ is a constant $f' \neq f$ and the predictions are continuously shifted due to the frequency mismatch.
    
    \item \textbf{Periodic Acceleration (AC).} The model relies on its previous predictions for the periodicity information and increases or decreases its belief $f'$ about the ground-truth frequency $f$ periodically in discrete increments.
\end{itemize}
We perform systematic evaluation with these defects in mind in \Cref{section:evaluation}.

%%%%%%%%%%%%%%%%%%%%%%%%%%%%%%%%%%%%%%%%%%%%%%%%%%%%%%%%%%%%%%%%%%%%%%%%%%%%%%%%%%%%%%%%%%%%%%%%%%%%%%%%%%%%%%%%%%%%%%%%%%%%%%%%%
\section{Population-Based Training for Periodic Generalisation}
\label{section:pbt}
%%%%%%%%%%%%%%%%%%%%%%%%%%%%%%%%%%%%%%%%%%%%%%%%%%%%%%%%%%%%%%%%%%%%%%%%%%%%%%%%%%%%%%%%%%%%%%%%%%%%%%%%%%%%%%%%%%%%%%%%%%%%%%%%%
To give a reasonable expectation on the possible performance of machine-learning models on the \textsc{PerKit}'s benchmark datasets we also evaluate the extrapolation performance of populations of simple models trained with Bayesian and genetic approaches.
In contrast to the recurrent networks and feedforward networks with periodic activations of \Cref{section:generalisationOverview}, populations of simple models benefit from being able to make multiple simultaneous guesses about the period of the signal being learned.
Further, once a ``good guess'' has been identified, they can start to exploit the guess to train ever better-performing models -- a behaviour hard to induce in traditional neural network training.

We consider three population-based models, \textsc{Bayes}, \textsc{n-Fittest}, and \textsc{Pareto}, for learning periodic signals with the objective to extrapolate.
These are all parametrised the same and train populations of networks of the same architecture.

\textsc{Bayes} performs Bayesian optimisation \cite{frazier2018tutorial}, aiming to minimise individual final training losses.
It does so by tuning its period guess by guided sampling in each iteration of the optimisation process.
\textsc{n-Fittest} and \textsc{Pareto} follow a common skeleton evolutionary algorithm with parametric neighbourhood crossovers, differing only in the method by which they choose the subset of the population that is to reproduce.
They too aim to minimise individual final training losses, but they do so by choosing the best-performing models to reproduce for each generation.
For simplicity and consistency, we will refer to sampling in \textsc{Bayes} as reproduction and to iterations of \textsc{Bayes} as generations.

The inputs to each of the above models are the assumed signal master period range $[r_a, r_b]$, root (starting) population $n_r$, and minimum number of individuals to reproduce each generation $n_g \leq n_r$.
Further, \textsc{n-Fittest} and \textsc{Pareto} also need a minimum number of descendants of different roots that the algorithm is to \textit{enforce to be present} among the population that is to reproduce -- $n_e \leq n_r$.
The algorithms run until the maximum number of generations has been reached or until a fitness threshold has been crossed.
Their output is a population $\mathcal{P}$, whose fittest individuals are those who are likely to achieve the best performance in terms of their ability to extrapolate pure periodic signals or periodic signals with trends.

\subsection{Individuals}
\begin{figure}[t]
    \centering
    \includegraphics[width=0.25\textwidth]{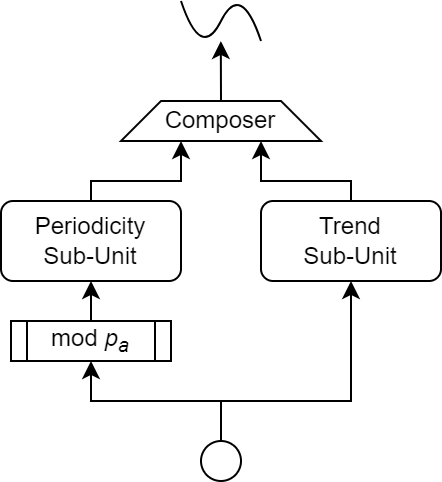}
    \caption{
        The architecture of a single population unit $a$. $p_a$ denotes the genetic parameter.
    }
    \label{fig:architecture}
\end{figure}

The individuals of the population, or population \textit{units}, are neural networks consisting of \textit{trend}, \textit{periodicity}, and \textit{composer} sub-units (\Cref{fig:architecture}).
In our experiments, we chose the trend sub-unit to be a linear feedforward network, periodicity sub-unit to be a feedforward network with ReLU-activated hidden layers and a linear output layer, and the composer to be a simple single-layer linear network.
We have also successfully experimented with a polynomial neuron (such as the one seen in GMDH networks) as the trend unit and aim to report on our findings in our future work.
Every individual $a$ has an associated genetic parameter $p_a$ representing its period estimate.
The input to population units is the time-component $x$ of a potentially periodic signal. $x$ is wired directly to the trend sub-unit, $x\mod p_a$ (modulo taken in direction from $-\infty$ to $\infty$) is fed into the periodicity sub-unit, and the outputs of the periodicity and trend sub-units are then forwarded to the composer, which yields the signal estimate $y_a$ of $a$.
This configuration assumes that a new period of the target signal begins at the origin ($0$).

The initial population consists of $n_r$ root individuals, with parameters $p_\bullet$ spaced evenly on $[r_a, r_b]$ including at the endpoints of the interval.
In our experiments, the weights of roots' sub-units are initialised with the Glorot uniform distribution \cite{glorot2010understanding}.
When individuals $a_1,a_2$ are chosen to reproduce, we designate the root ancestor of the fitter of the two roots the root ancestor of their offspring.
The root ancestor of each root is the root itself.

We number generations as $g=1,2,\dotsc$ (the zeroth generation is the root population).
For each generation, we first train previously untrained units on 80\% of the available training data with mean squared error loss, and use the remaining data for validation.
We terminate training early if the validation loss stops improving for a number of epochs.

\subsection{Evolution of \textsc{Bayes}}
We simply follow the iterative procedure for Bayesian optimisation \cite{frazier2018tutorial}.

\subsection{Evolution of \textsc{Pareto}, \textsc{n-Fittest}}
While it could be argued that the fitness should be assessed on the basis of an evaluation sample that is taken from outside the training range, it is precisely the point that our models learn to extrapolate periodically without any information about the signal besides what is available in training.
For each individual $a$ we keep the end validation loss $\ell_a$ (the ``unfitness'' of $a$). 

Once the training phase has been completed, we proceed with selecting the set of ``best'' candidates for reproduction $\mathcal{B}$.
For the \textsc{n-Fittest} algorithm, we use $-\ell_a$ as the measure of fitness and take the $n_g$ individuals with least $\ell_\bullet$ to be $\mathcal{B}$, choosing the younger individual in case of a tie.
For the \textsc{Pareto} algorithm, we fit a Pareto distribution across the range of losses in $\mathcal{P}$ and calculate Pareto fitness scores $s_\bullet$ according to the formula

\begin{align*}
    \mu_a &= \frac{\ell_a - \min_{z \in \mathcal{P}} \ell_z }{\max_{z \in \mathcal{P}} \ell_z - \min_{z \in \mathcal{P}} \ell_z} \,\,\,\,
    s_a &= \frac{S\sqrt{g}}{ (1+\mu_a)^{1 + S\sqrt{g}} },
\end{align*}

where $S$ is a score-scaling hyperparameter of the model controlling the exploration-exploitation balance and $\mu_\bullet$ is the validation loss normalised to the range seen in $\mathcal{P}$.
$n_g$ candidates for reproduction are then chosen to form $\mathcal{B}$ according to a multinoulli distribution where the probability of individual $a$ being selected is proportional to $s_a$.

We then count the number of distinct root ancestors $d_r$ among candidates in $\mathcal{B}$ and if it is less than $n_e$, we keep adding least unfit individuals not yet in $\mathcal{B}$ with root ancestors different from all the previous until $d_r=n_e$.

Finally, we perform the crossovers. For every $b_2 \in \mathcal{B}$, we choose the other parents $b_1,b_3$ such that 
\[
    b_1 = \argmax_{z \text{ s.t. } p_z<p_{b_2}} p_z \text{\, and \, }
    b_3 = \argmin_{z \text{ s.t. } p_z>p_{b_2}} p_z.
\]
We place parameter of the offspring individuals $b_1,b_2$ proportionally to the parents' fitness scores.
Let $b_i,b_j$ be parents such that $p_{b_i} < p_{b_j}$.
We first calculate the pair-relative fitness scores,
\begin{align*}
    \sigma^{\textsc{n-Fittest}}_{b_i} = \frac{\ell_{b_i}}{\ell_{b_i} + \ell_{b_j}} \hspace{20pt} & \sigma^{\textsc{n-Fittest}}_{b_j} = 1 - \sigma^{\textsc{n-Fittest}}_{b_i} \\
    \sigma^{\textsc{Pareto}}_{b_i} = \frac{s_{b_j}}{s_{b_i} + s_{b_j}} \hspace{20pt} & \sigma^{\textsc{Pareto}}_{b_j} = 1 - \sigma^{\textsc{Pareto}}_{b_i},
\end{align*}

(note the contrasting roles of $b_i$ and $b_j$ per case), and then compute the new parameter,
\[
    p_{c} = p_{b_i} + \sigma^{\bullet}_{b_i}(p_{b_j} - p_{b_i})
\]
We perform this procedure for the pairs $b_1,b_2$ and $b_2,b_3$ to get children parameters $p_{c_1},p_{c_2}$ respectively.

Instead of initialising units for children $c_i$ afresh, we clone the trained state of $b_2$ (central parent) into $c_i$ and begin the training from that state.
Experimenting, we learned that while this approach does not lead to consistent improvements in terms of accuracy of our models' predictions when compared to starting from default weight initialisation, it significantly accelerates the training.

The offspring of $\mathcal{B}$ are then added to $\mathcal{P}$ and if the pre-specified termination criteria such as accuracy or maximum number of generations are not met, the algorithms proceed with the next generation.

\begin{figure}[t]
    \centering
    \includegraphics[width=0.50\textwidth]{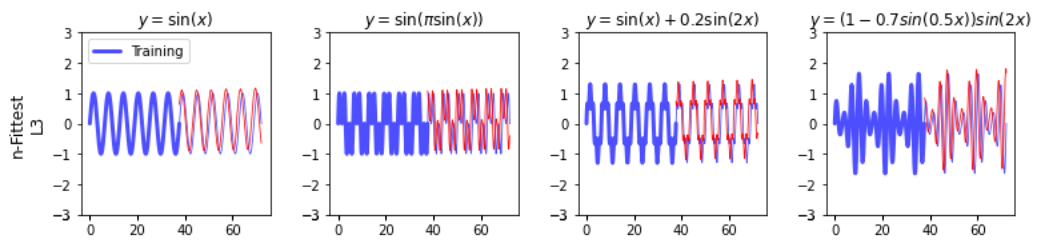}
    \caption{
        An example of the training and predictions of the \textsc{n-Fittest} model with configuration as in \Cref{fig:hierarchy}. $n_r = 8, n_g = 7, n_e = 3, r_a = 3\pi, r_b = 5\pi$.
    }
    \label{fig:nFittest}
\end{figure}

We illustrate the outputs of \textsc{n-Fittest} in \Cref{fig:nFittest}.

%%%%%%%%%%%%%%%%%%%%%%%%%%%%%%%%%%%%%%%%%%%%%%%%%%%%%%%%%%%%%%%%%%%%%%%%%%%%%%%%%%%%%%%%%%%%%%%%%%%%%%%%%%%%%%%%%%%%%%%%%%%%%%%%%
\section{Evaluation}
\label{section:evaluation}
%%%%%%%%%%%%%%%%%%%%%%%%%%%%%%%%%%%%%%%%%%%%%%%%%%%%%%%%%%%%%%%%%%%%%%%%%%%%%%%%%%%%%%%%%%%%%%%%%%%%%%%%%%%%%%%%%%%%%%%%%%%%%%%%%

\subsection{Method}
To systematically evaluate the ability of a neural model to generalise extrapolatively, we design a straightforward but rich collection of classes of continuous periodic signals with hierarchically increasing complexity.

We begin with generation of periodic forms.
Recursively, a periodic form is either an elementary form (see \Cref{table:forms}) or a sum, product, or chained application of a periodic form and an elementary form.
More formally,
\begin{align*}
    \textsc{Form}_0 &:= \textsc{Form}_E \\
    \textsc{Form}^{\, +\times\circ}_i &:= \{ f+g,fg,f \circ g \\ & \qquad \quad : f \in \textsc{Form}_E, g \in \textsc{Form}_{i-1} \}
\end{align*}
where $i \geq 1$, $\textsc{Form}_E$ is the set of elementary forms, $\textsc{Form}^{-}_i$ the set of forms of order $i \geq 0$ arrived at by combining lower-order forms by operations in superscript.

On top of learning to generalise in periodicity, some models are also able to capture a general trend offsetting an otherwise periodic signal.
To represent that in our signal collection, we consider polynomial and exponential trends.
Denoting the trend forms (\Cref{table:forms}) by $\textsc{Form}_T$, a general form is a sum of a periodic form and a trend form.

\subsection{Experiments}
\label{section:experiments}
We conduct three experiments, assessing the ability of models to extrapolate (L2-L3): periodic signals, periodic signals with noisy training, and signals generated by offsetting a periodic base with a linear trend.

For the noiseless and noisy periodic signals we use forms in $\textsc{Form}^{\, +\times}_2$, for signals with trend we use forms $\{f + g : f \in \textsc{Form}^{\, +\times}_2, g \in \textsc{Form}_T \}$.
For each form we generate $n_v$ variants, drawing every coefficient including frequencies uniformly at random from a fixed range.
Then, we choose a master period $\tau$ from $[0.5,1.0]$ uniformly at random and normalise individual frequencies so that the variant is a periodic signal with period $\tau$.
The training domain is $\mathcal{D}_T = (-n_T \tau, n_T \tau)$ and we evaluate on $\mathcal{D}_E = (-n_E \tau, n_E \tau) \backslash \mathcal{D}_T$, where $n_T, n_E$ represent the numbers of periods to train/evaluate on.
The signal is normalised by a constant so that all values on $\mathcal{D}_E$ lie between $-1$ and $1$ inclusive.
In the noisy scenario, we further add Gaussian noise with mean $0$ and variance $\sigma^2$.
The data is then sampled uniformly at random and used for training.
We repeat the training and evaluation independently $n_r$ times for each variant.

\begin{table}[]
\caption{
    An overview of forms used to evaluate extrapolative generalisation abilities.
    \textit{square}, \textit{saw}, and \textit{poly} denote the unit square wave with 50\% duty cycle, unit positive sawtooth wave, and symmetric bilateral polynomial wave of order $n$, respectively, each with period $1$.
}
\centering
\begin{tabular}{@{}lllll@{}}
\toprule
Name                & Type  & Analytic Form &  \\
\midrule
square wave         & elementary     & $\text{square}(\frac xT + \phi)$      &  \\
sawtooth wave       & elementary     & $\text{saw}(\frac xT + \phi)$         &  \\
sinusoid            & elementary     & $\sin(2\pi\frac xT + 2\phi\pi)$       &  \\
tangent             & elementary     & $\tan(\pi\frac xT + \phi\pi)$         &  \\
polynomial wave of order $n$        & elementary     & $\text{poly}(\pi\frac xT + \phi\pi, n)$   &  \\
polynomial trend of order $n$       & trend     & $c_0 + c_1 x + \dotsc + c_n x^n$          &  \\
exponential trend                   & trend     & $c_0 e^{c_1 x}$                           &  \\
\bottomrule
\end{tabular}
\label{table:forms}
\end{table}

The literature on trigonometric activation functions \cite{parascandolo2016taming,zhumekenov2019fourier} cites problems with their optimisation due to the existence of a series of local minima.
In an attempt to counter this, we ran all of the above models on the whole dataset $6$ times -- once for each optimiser among SGD, RMSprop, Adam, AdaMax, AdaDelta, and Nadam.
We compared the aggregate results for the whole benchmarking dataset and observed no significant differences between the results of different optimisers.
Our final results, reported below, are thus computed on the collation of runs, disregarding the optimiser used.

We chose a uniform width of 64 hidden units for all feedforward networks, including those used by \textsc{n-Fittest} and \textsc{Pareto}.
In the case of snake activation functions and by the analysis of \cite{ziyin2020neural} this would correspond to a target of 32 (not necessarily consecutive) harmonics in the approximating Fourier series, while for $\sin + \cos$ \cite{liu2013fourier} this corresponds to 64 harmonics.
Individual experiments showed that adding more hidden neurons did not improve the extrapolative performance of feedforward networks.
We also set the number of recurrent units to 64 for LSTM, GRU, and simple RNNs, but note that recurrent units are more complex in their structure than individual feedforward neurons.
For the population units of neuroevolutionary models, we use a two-layer periodic unit 60 neurons wide with ReLU activation on the hidden layer, and a trend unit consisting of a single linear neuron.

\vspace{-1pt}
\subsection{Metrics}
\label{section:metrics}
As was shown in \Cref{section:generalisationOverview}, neural models might exhibit a wide range of behaviours in periodic extrapolation scenarios, and straightforward per-point losses such as mean squared error (MSE) or mean absolute error (MAE) may not be representative of the models' ability to learn information pertaining to signal periodicity and to extrapolate beyond the training range.

We therefore use a range of metrics to capture and report on prevalence of phenomena seen in \Cref{section:generalisationOverview} such as periodic signal shift, signal frequency shift (signal ``speed-up''), and signal frequency acceleration.
Each of the evaluation metrics is based on a point-based metric $\mathpzc{m}$ and is arrived at by minimising $\mathpzc{m}$ across a parameter trying to correct the predictions.
Let $x,y(x),y'(x)$ be the position, true, and predicted signal values respectively, $e_T$ the nearest endpoint of the training domain, and $\tau$ a period of the signal.
If the signal is offset by a trend or otherwise modified, $\tau$ is a period of the periodic component.
We then define
\begin{align*}
    \varA{SH\mathpzc{m}} & = \min_{w \in (-\epsilon, +\epsilon)} \mathpzc{m}\left(x, y(x), y'\left(x+w\left\lfloor \frac{x-e_T}{\tau} \right\rfloor\right)\right) \\
    \varA{SP\mathpzc{m}} & = \min_{w \in (-\epsilon, +\epsilon)} \mathpzc{m}\left(x, y(x), y'\left(x(1 + w)\right)\right) \\
    \varA{AC\mathpzc{m}} & = \min_{w \in (-\epsilon, +\epsilon)} \mathpzc{m}\left(x, y(x), y'\left(x\left(1 + w\left\lfloor \frac{x-e_T}{\tau} \right)\right\rfloor\right)\right)
\end{align*}

The choice of the metrics specific to degradation phenomena can be justified as assuming that the corresponding prediction degradation phenomenon is present and finding the minimum value of point-based metric $\mathpzc{m}$ for a parameter characterising the decay in predictive ability within some permissible range.

Further, in our experiments we observed that the tails of the extrapolative predictions (i.e. the segments of the signal farthest from the training domain) often disproportionally affected the summary metrics, and in response we chose to weigh the point contributions to the original metrics decreasingly with the increasing distance from the training domain.
Let $\mathpzc{m}$ be a point-based metric.
The the distance-adjusted $\mathpzc{m}$ is
\[
    \varA{DA-\mathpzc{m}}(x, y, y') = \mathpzc{m}(y, y')\left(\frac{ x - e_T }{ d_T }\right)^\alpha
\]
where $d_T$ is the diameter of the training domain and $\alpha \geq 1$ is the weighing decay parameter.
The comparison of the values for a per-point metric $\mathpzc{m}$ and the distance-adjusted $\mathpzc{m}$ allows us to quickly assess the extent to which the predictions of the model deteriorate with increasing distance from the training domain. 

\subsection{Results}
\label{section:results}

The results of our experiments are shown in \Cref{table:noiselessResultsMeanStddev,table:noisyResultsMeanStddev,table:trendResultsMeanStddev}.
We performed the experiments with $n_v = 10, n_r = 5, n_T = 5, n_E = 10, \sigma^2 = 0, \alpha = 1$, sampling rate 100 per period, and recurrent window of length $7$.
PBT methods were run for exactly 10 generations with $n_r = 8, n_g = 7, n_e = 3, r_a = 0.5, r_b = 1.0$.
Each experimental instance was repeated $6$ times, once for each optimiser considered.
For metric evaluation, $\epsilon = 0.05$, and 21 samples were taken over $[-\epsilon, \epsilon]$.
\textbf{Emph.} and \color{BlueViolet}{emph.} \color{Black}{denote} the best performance per metric and model, respectively.

In the scenario with noiseless periodic signals (\Cref{table:noiselessResultsMeanStddev}), we observe that the snake activations outperform $x+\sin$ and $x+\cos$, and that the snake feedforward neural network with frozen frequency parameters further outperforms snake networks with trainable frequencies.
All of the recurrent networks outperform the feedforward networks, and our genetic models \textsc{n-Fittest} and \textsc{Pareto} further improve on the best-performing recurrent networks by 32-35\% in MSE and 45-46\% in SHDA-MSE.

There was a noticeable drop in performance of feedforward networks when noisy signals were considered (\Cref{table:noisyResultsMeanStddev}).
Both recurrent and PTB models remained largely unaffected.

On trend-offset periodic signals (\Cref{table:trendResultsMeanStddev}), we observe that $x+\sin$ and $x+\cos$ show performance comparable to snake activations and that GRU networks outperform simple recurrent networks and LSTM architectures.
It is surprising that GRU networks perform very well on the trend-offset data, tentatively suggesting that the gated recurrent unit possesses some ability to preserve shape while recognising the presence of a linear trend.
The genetic models \textsc{n-Fittest} and \textsc{Pareto} further improve on the best-performing recurrent networks by 77-79\% in MSE and ~85\% in SHDA-MSE and SPDA-MSE.

%%%%%%%%%%%%%%%%%%%%%%%%%%%%%%%%%%%%%%%%%%%%%%%%%%%%%%%%%%%%%%%%%%%%%%%%%%%%
% NOISELESS MEANS
%%%%%%%%%%%%%%%%%%%%%%%%%%%%%%%%%%%%%%%%%%%%%%%%%%%%%%%%%%%%%%%%%%%%%%%%%%%%
\begin{table}[t]
\caption{
    Results of the experiments on noiseless periodic signals.
}
\centering
\scalebox{1}{
\begin{tabular}{@{}lcc ccc@{}}
    \toprule
     & MSE & DA- & SHDA- & SPDA- & ACDA- \\ \midrule
    $\sin$           & 0.209    & 0.149    & 0.148    & \color{BlueViolet}{0.147}  & 0.147 \\
    $\cos$ \cite{silvescu1999fourier}           & 0.245    & 0.174      & 0.172   & \color{BlueViolet}{0.171}       & 0.171\\
    $\sin + \cos$ \cite{liu2013fourier}         & 0.319    & 0.220      & 0.218   & \color{BlueViolet}{0.217}      & 0.217 \\
    $x+\sin$ \cite{ziyin2020neural}             & 5.095    & 3.624      & 3.620   & \color{BlueViolet}{0.317}      & 3.621 \\
    $x+\cos$ \cite{ziyin2020neural}             & 4.662    & 3.289      & 3.283   & 3.281       & \color{BlueViolet}{3.280} \\
    snake \cite{ziyin2020neural}                & 0.375    & 0.264      & 0.262   & \color{BlueViolet}{0.261}       & 0.261 \\
    t-snake \cite{ziyin2020neural}              & 0.391    & 0.275      & 0.273   & \color{BlueViolet}{0.272}       & 0.272 \\
    SRNN                                        & 0.081    & 0.811      & 0.074   & \color{BlueViolet}{0.070}       & 0.070 \\
    GRU                                         & 0.071    & 0.072      & 0.065   & \color{BlueViolet}{0.061}       & 0.062 \\
    LSTM                                        & 0.076    & 0.076      & 0.069   & \color{BlueViolet}{0.063}       & 0.065 \\
    
    \midrule
    \textsc{Bayes}             & 0.050    & 0.038      & 0.037   & \color{BlueViolet}{0.037}       & 0.049 \\
    \textsc{n-Fittest}          & 0.048    & 0.037      & \color{BlueViolet}{0.036}   & 0.036       & \textbf{0.048} \\
    \textsc{Pareto}             & \textbf{0.046}    & \textbf{0.036}      & \color{BlueViolet}{\textbf{0.035}}   & \textbf{0.035}       & 0.049 \\
    \bottomrule
    \end{tabular}
}
\label{table:noiselessResultsMeanStddev}
\end{table}

%%%%%%%%%%%%%%%%%%%%%%%%%%%%%%%%%%%%%%%%%%%%%%%%%%%%%%%%%%%%%%%%%%%%%%%%%%%%
% NOISY MEANS
%%%%%%%%%%%%%%%%%%%%%%%%%%%%%%%%%%%%%%%%%%%%%%%%%%%%%%%%%%%%%%%%%%%%%%%%%%%%
\begin{table}[t!]
\caption{
    Results of the experiments on noisy periodic signals with $\sigma^2 = 0.15$.
}
\centering
\scalebox{1}{
\begin{tabular}{@{}lcc ccc@{}}
    \toprule
     & MSE & DA- & SHDA- & SPDA- & ACDA- \\ \midrule
    $\sin$            & 0.338    & 0.232    & 0.231    & \color{BlueViolet}{0.228}  & 0.274 \\
    $\cos$ \cite{silvescu1999fourier}           & 0.373    & 0.274      & 0.273   & \color{BlueViolet}{0.269}           & 0.322 \\
    $\sin + \cos$ \cite{liu2013fourier}         & 0.358    & 0.250      & 0.248   & \color{BlueViolet}{0.248}           & 0.247 \\
    $x+\sin$ \cite{ziyin2020neural}             & 0.399    & 0.279      & 0.277   & \color{BlueViolet}{0.276}           & 0.276 \\
    $x+\cos$ \cite{ziyin2020neural}             & 0.257    & 0.181      & 0.180   & \color{BlueViolet}{0.179}           & 0.179 \\
    snake \cite{ziyin2020neural}                & 0.391    & 0.274      & 0.272   & \color{BlueViolet}{0.271}           & 0.271 \\
    t-snake \cite{ziyin2020neural}              & 0.435    & 0.305      & 0.303   & \color{BlueViolet}{0.302}           & 0.302 \\
    SRNN                                        & 0.075    & 0.075      & 0.069   & \color{BlueViolet}{0.065}           & 0.065 \\
    GRU                                         & 0.072    & 0.072      & 0.065   & \color{BlueViolet}{0.059}           & 0.061 \\
    LSTM                                        & 0.073    & 0.073      & 0.066   & \color{BlueViolet}{0.061}      & 0.063 \\
    \midrule
    \textsc{Bayes}             & 0.051    & 0.039      & 0.038   & \color{BlueViolet}{0.038}       & 0.049 \\
    \textsc{n-Fittest}                          & \textbf{0.045}    & 0.035      & \color{BlueViolet}{\textbf{0.034}}   & \textbf{0.034}       & \textbf{0.049} \\
    \textsc{Pareto}                             & 0.046    & \textbf{0.035}      & \color{BlueViolet}{0.035}   & 0.035       & 0.049 \\
    \bottomrule
    \end{tabular}
}
\label{table:noisyResultsMeanStddev}
\end{table}

%%%%%%%%%%%%%%%%%%%%%%%%%%%%%%%%%%%%%%%%%%%%%%%%%%%%%%%%%%%%%%%%%%%%%%%%%%%%
% TREND MEAN/STDDEV
%%%%%%%%%%%%%%%%%%%%%%%%%%%%%%%%%%%%%%%%%%%%%%%%%%%%%%%%%%%%%%%%%%%%%%%%%%%%
\begin{table}[t!]
\caption{
    Results of the experiment on trend-offset periodic signals without noise.
}
\centering
\scalebox{1}{
\begin{tabular}{@{}lcc ccc@{}}
    \toprule
     & MSE & DA- & SHDA- & SPDA- & ACDA- \\ \midrule
    $\sin$         & 0.338    & 0.232    & 0.231    & \color{BlueViolet}{0.228}  & 0.274 \\
    $\cos$ \cite{silvescu1999fourier}           & 0.373    & 0.274      & 0.273   & \color{BlueViolet}{0.269}      & 0.322 \\
    $\sin + \cos$ \cite{liu2013fourier}         & 0.520    & 0.371      & 0.370   & \color{BlueViolet}{0.367}      & 0.431 \\
    $x+\sin$ \cite{ziyin2020neural}             & 0.486    & 0.336      & 0.334   & \color{BlueViolet}{0.331}      & 0.432 \\
    $x+\cos$ \cite{ziyin2020neural}             & 0.342    & 0.237      & 0.236   & \color{BlueViolet}{0.235}      & 0.346 \\
    snake \cite{ziyin2020neural}                & 0.441    & 0.307      & 0.305   & \color{BlueViolet}{0.303}      & 0.381 \\
    t-snake \cite{ziyin2020neural}              & 0.498    & 0.345      & 0.344   & \color{BlueViolet}{0.342}      & 0.451 \\
    SRNN                                        & 0.081    & 0.081      & 0.074   & \color{BlueViolet}{0.070}      & 0.070 \\
    GRU                                         & 0.026    & 0.026      & 0.019   & \color{BlueViolet}{0.018}      & 0.034 \\
    LSTM                                        & 0.366    & 0.366      & 0.355   & \color{BlueViolet}{0.351}      & 0.378 \\
    \midrule
    \textsc{Bayes}                                       & 0.006    & 0.005      & \color{BlueViolet}{0.004}   & 0.004       & 0.107 \\
    \textsc{n-Fittest}              & 0.006    & 0.005      & \color{BlueViolet}{0.004}   & 0.004      & 0.107 \\
    \textsc{Pareto}                 & \textbf{0.006}    & \textbf{0.005}      & \color{BlueViolet}{\textbf{0.004}}   & \textbf{0.004}       & \textbf{0.107} \\
    \bottomrule
    \end{tabular}
}
\label{table:trendResultsMeanStddev}
\end{table}

Comparing MSE with DA-MSE allows us to judge whether there are large prediction deviations from the target signal at the points far from the training range.
We observed that the gap (15-33\%) between MSE and DA-MSE was particularly common for trigonometric and snake feedforward networks, irrespective of whether they used activations monotonicitised by linear offset.
There was hardly any difference between MSE and DA-MSE for for recurrent and evolutionary models.

We see that the most common mode of extrapolative signal degradation is that of periodic speedup (cf. \Cref{section:modesOfDegradation}), and that the difference is particularly noticeable (in terms of relative improvement over base metrics) in recurrent neural networks. 

We attribute the generally comparable performance of \textsc{n-Fittest} and \textsc{Pareto} to the simplicity of the search space and note that a significant difference is immediately seen if the assumed master period range $[r_a, r_b]$ is made larger and contains several multiples of the master period.
We also note from the results that trigonometric activations that had been proposed for L2 tasks do not perform well on L3 tasks, consistently with our outline of the difficulty hierarchy for periodicity learning.

%%%%%%%%%%%%%%%%%%%%%
\subsection{Training Resource Consumption of PBT Methods}
%%%%%%%%%%%%%%%%%%%%%
Pupulation-based training and especially evolutionary methods are often associated with high computational cost and long program runtime.

We trained and evaluated our models on a single NVIDIA TITAN Xp GPU.
For the performance comparison, we trained all our models with the Adam optimiser \cite{kingma2014adam}.
Thanks to the simplicity the genetic unit architecture, the full multi-generational evolutionary training and evaluation of \textsc{Bayes}, \textsc{n-Fittest}, and \textsc{Pareto} took only 12-15\% longer than that of snake feedforward neural networks with frozen frequencies and 40-44\% shorter than snake networks with trainable frequency parameters, while outperforming both in terms of the metrics above in all scenarios.

This suggest that population-based training, albeit naive in its motivation, currently trumps the best available models tailored for periodic extrapolation in practice. 

%%%%%%%%%%%%%%%%%%%%%%%%%%%%%%%%%%%%%%%%%%%%%%%%%%%%%%%%%%%%%%%%%%%%%%%%%%%%%%%%%%%%%%%%%%%%%%%%%%%%%%%%%%%%%%%%%%%%%%%%%%%%%%%%%
\section{Conclusion}
\label{section:conclusion}
%%%%%%%%%%%%%%%%%%%%%%%%%%%%%%%%%%%%%%%%%%%%%%%%%%%%%%%%%%%%%%%%%%%%%%%%%%%%%%%%%%%%%%%%%%%%%%%%%%%%%%%%%%%%%%%%%%%%%%%%%%%%%%%%%
We have identified periodic extrapolation as the computationally simplest mode of extrapolative generalisation. 
Our work systematically evaluates network architectures thought to generalise well beyond training domains and finds that traditional recurrent neural networks outperform all of the architectures proposed specifically to tackle extrapolation. Moreover, the latest invention proposed -- snake activation -- also consistently fails at periodic extrapolation and offers little to no improvement over purely periodic activations.
We believe that these results suggest that there does not yet exist a universal architecture with plausible extrapolative properties, despite the claims of effectiveness founded on the existence of weights corresponding to the Fourier series of the target signal.

With the classical Bayesian and neuroevolutionary population-based training methods outperforming all other models while keeping their training time below that of the most recent architectures for periodic generalisation, we view their performance as natural baselines to be overcome by future work.

All our code is available as \textsc{PerKit}\footref{footnote:repo_link} -- a toolkit for the study of periodicity in neural networks.
\textsc{PerKit} has been designed specifically to allow making of future benchmarking extensions and additions of new models for evaluation with ease.
We hope that together with our study it will help to facilitate research in extrapolative generalisation.

%%%%%%%%%%%%%%%%%%%%%%%%%%%%%%%%%%%%%%%%%%%%%%%%%%%%%%%%%%%%%%%%%%%%%%%%%%%%%%%%%%%%%%%%%%%%%%%%%%%%%%%%%%%%%%%%%%%%%%%%%%%%%%%%%
\bibliographystyle{plain} % We choose the "plain" reference style
\bibliography{bibliography} % Entries are in the refs.bib file
%%%%%%%%%%%%%%%%%%%%%%%%%%%%%%%%%%%%%%%%%%%%%%%%%%%%%%%%%%%%%%%%%%%%%%%%%%%%%%%%%%%%%%%%%%%%%%%%%%%%%%%%%%%%%%%%%%%%%%%%%%%%%%%%%

\end{document}